\documentclass{article}
\usepackage{spconf,amsmath,graphicx}

\usepackage{url,amsfonts,enumitem,booktabs,subfigure,multirow}
\usepackage{algorithm}
\usepackage{algorithmic}

\title{Predict and Interpret Health Risk using EHR through Typical Patients}

\name{Zhihao Yu$^{1,2}$, Chaohe Zhang$^{1,2}$, Yasha Wang$^{2,3,*}$, Wen Tang$^{4}$, Jiangtao Wang$^{5}$, Liantao Ma$^{2,3,*}$\thanks{*Corresponding author.}}
\address{$^{1}$School of Computer Science, Peking University, Beijing, China\\
$^{2}$Key Laboratory of High Confidence Software Technologies, Ministry of Education, Beijing, China\\
$^{3}$National Engineering Research Center of Software Engineering, Peking University, Beijing, China\\
$^{4}$Division of Nephrology, Peking University Third Hospital, Beijing, China\\
$^{5}$The Centre for Intelligent Healthcare, Coventry University, UK\\
}
\begin{document}
\ninept
\maketitle
\begin{abstract}
Predicting health risks from electronic health records (EHR) is a topic of recent interest. Deep learning models have achieved success by modeling temporal and feature interaction. However, these methods learn insufficient representations and lead to poor performance when it comes to patients with few visits or sparse records. 
Inspired by the fact that doctors may compare the patient with typical patients and make decisions from similar cases, we propose a Progressive Prototypical Network (PPN) to select typical patients as prototypes and utilize their information to enhance the representation of the given patient. 
In particular, a progressive prototype memory and two prototype separation losses are proposed to update prototypes. Besides, a novel integration is introduced for better fusing information from patients and prototypes.
Experiments on three real-world datasets demonstrate that our model brings improvement on all metrics. To make our results better understood by physicians, we developed an application at \url{http://ppn.ai-care.top}. Our code is released at \url{https://github.com/yzhHoward/PPN}.
% In short, PPN outperform existing health risk prediction methods while giving interpretations.
\end{abstract}
\section{Introduction}
As electronic systems widely used in hospitals, electronic health records (EHR) can be used to build models to predict health risks and improve healthcare quality.
In recent years, deep learning approaches have achieved early success using EHR data. Most of these works attempt to model temporal information \cite{song2018attend,ma2020concare,ma2020adacare}, capture feature correlation \cite{ma2020concare,choi2016retain}, and find important visits \cite{ma2017dipole,gao2020stagenet,bai2018interpretable}.

However, patients with few visits or highly sparse records are common in EHR data. For example, some patients may visit multiple hospitals, resulting in some records being unavailable due to privacy protection \cite{miller2009privacy}; or not testing the same indicators in adjacent visits, resulting in sparse observations.
% The indicators in the datasets may differ, introducing inevitable missing observations.
In these cases, traditional representation learning methods may only capture insufficient information and learn low-discriminative representations. 
% Though works are proposed to contend with missing observation \cite{ma2020adacare,gao2020stagenet}, patients with few visits have not been fully investigated in these approaches.
% such as Ma et al.\cite{ma2020adacare} adopting dilated convolution and Gao et al. \cite{gao2020stagenet} applying stage-aware RNN to explore long-term information. Moreover, works based on generative methods, such as Shukla et al. \cite{shukla2021multi} and Mulyadi et al. \cite{mulyadi2021uncertainty} try to impute missingness. 
% Nevertheless, patients with few visits have not been fully investigated in these approaches.

To tackle this problem, we draw inspiration from the routine practices of doctors: When a doctor receives a new patient, the doctor may compare the patient with previous typical cases to understand the health status \cite{groopman2007doctors,arrieta2020explainable,raghu2018diagnosis}. 
% In addition, clinical practice guidelines \cite{raghu2018diagnosis} provide typical patients for doctors to learn diagnosis skills, and studies \cite{striano2008typical} adopt typical patients to help doctors make treatment decisions.
% The doctor may not understand the situation well for a patient with only a few visits.
% However, supposing the doctor is told that the patient is similar to a typical patient, the doctor is likely to make a more accurate judgment since the typical patient provides valuable information. Web-based EHR provides opportunity for users to observe typical patients.
\textit{So, can we select typical patients and utilize their knowledge to enhance representations of patients?}
We attempt to build a model to select typical cases as prototypes and facilitate the representations of patients through these cases. 
% \rev{3: this paragraph is wordy. try to simplify and abstract into several sentences}
% However, there are two open challenges to be solved. 

While learning and utilizing typical patients, there are following challenges. One the one hand, the physical status of the patient is complicated. There are many implicit subtypes among patients, and the patients belonging to the same subtype may have different outcomes \cite{anders2018ckd}. Learning prototypes by outcomes like traditional methods \cite{snell2017prototypical,chen2018looks} ignores the subtypes and introduces bias. Besides, the clustering structure between prototypes and patients is crucial. If only use cluster centroids as prototypes at the beginning of the training, the cluster structure cannot be maintained as model updates. Some prototypes may collapse into a single point while training, thereby losing diversity and confusing doctors. On the other hand, how to use prototypes for prediction is still a challenging problem. Most works \cite{xu2020attribute,doras2020prototypical} perform prediction using similarity or distance between prototypes and input, but these manners omit the original information from input. In health prediction, the information from patients is essential and the similarity alone is insufficient.

In this paper, we introduce prototype learning into EHR analysis on the health risk prediction and propose a Progressive Prototypical Network (PPN). PPN learns patient representation by multi-channel feature extraction, and then enhance it by prototypical feature integration with typical patients for health prediction. In selecting typical patients, PPN selects them as prototypes by novel progressive method and utilze prototype separation losses to promote diversity.
PPN is enabled by the following contributions:
\begin{itemize}[leftmargin=*]
\setlength{\itemsep}{0pt}
\setlength{\parsep}{0pt}
\setlength{\parskip}{0pt}
% \setlength{\topsep}{0pt}
% \rev{8: you'd better to describe what task has been performed in your paper, rather than the detail method in the first contribution item}
\item A progressive prototype memory and two prototype separation losses and are proposed to obtain typical patients perceptively while ensuring cluster structure.
\item An integration method of prototype features is proposed to compute the similarity between prototypes and patients and then incorporate their information.
\item Experiments on three datasets verify the effectiveness of PPN. Especially, PPN performs well on data with high missing rate.
% All findings are in accord with medical experts and literature.
\item An interactive application is built to visualize the prediction and interpretation from PPN at \url{http://ppn.ai-care.top}.
\end{itemize}

\begin{figure*}[h]
  \centering
  \includegraphics[width=1.8\columnwidth]{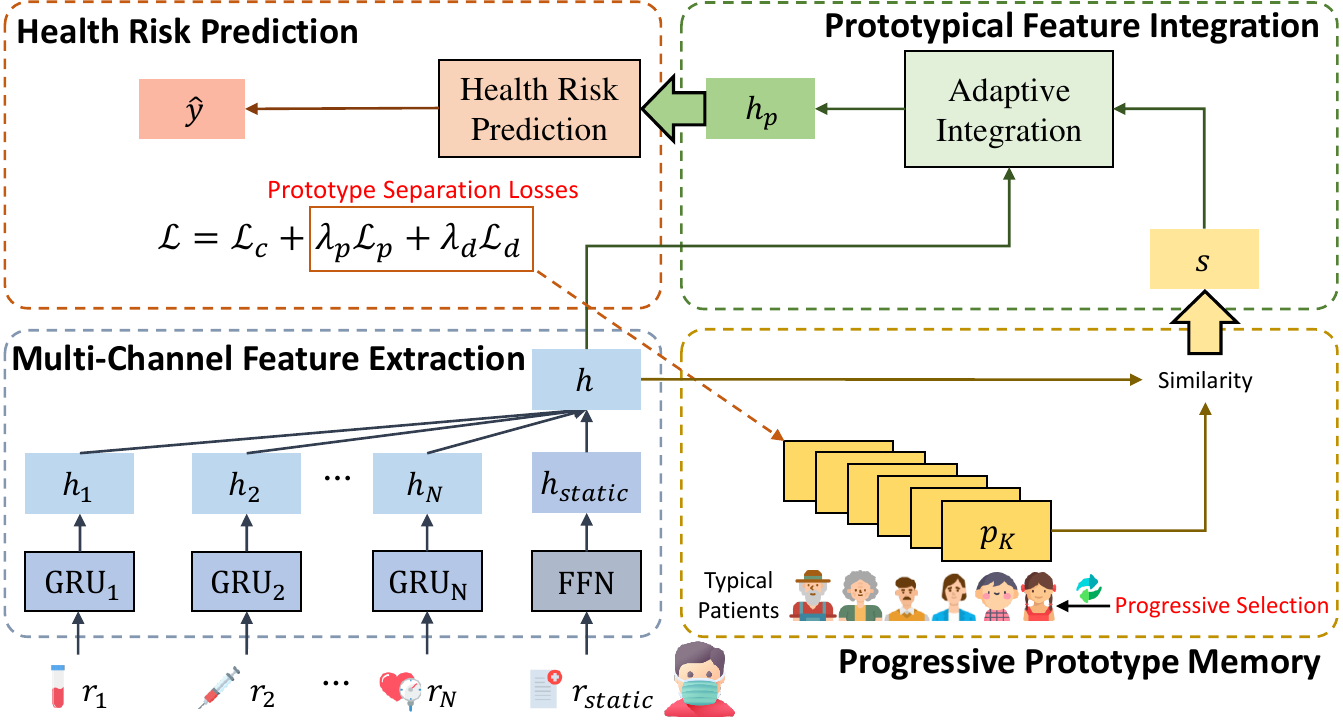}
  \caption{The framework of PPN. For the EHR of a patient, PPN firstly extracts its health status, computes the similarity coefficient between typical patients, re-weight the health status, and predicts the health risk via the integrated feature}
  \label{framework}
\end{figure*}

\section{Preliminary}
Formally, EHR data can be denoted by $(\mathbb{X}, \mathbb{Y})$, where $\mathbb{X} \subseteq \mathbb{R}^{T \times N} \cup \mathbb{R}^{M}$ is the input space and $\mathbb{Y} = \{0, 1\}$ is the output space. The EHR records $x \in \mathbb{X}$ of a patient is defined as: given a series of $N$-dimensional laboratory tests and dynamic diagnoses, then the record of $n$-th indicators can be represented by $r_n$. For different patients, the lengths of series can be different.
Additionally, patients have $M$-dimensional demographic data and static diagnoses formulated as $r_{static}$.
The health risk prediction problem is to predict whether death risk events occur in the specified time window.

\section{Methodology}

\subsection{Overview}
As depicted in Figure \ref{framework}, PPN can be divided into four parts: multi-channel feature extraction module, progressive prototype memory, prototypical feature integration, and health risk prediction. In the following, each module will be described in detail.

% \begin{itemize}[leftmargin=*]

% \item The \textit{multi-channel representation extracting module} is developed to extract information from the input longitudinal patient matrix and static data to obtain the latent health status representations.

% \item The \textit{patient similarity modeling module} can calculate the similarity of health status representation between the patients and the learned prototypes.
% This module employs the health status of patients as prototypes. The prototype selection algorithm will be executed to acquire better prototypes.
% % \tbd{3}
% % 既然这几个module的叙述层次是并列的，那么描述句法也并列。都有一般现在时主动语态。不要上面是被动，下面有变成can do。就统一does sth

% \item The \textit{prototype attention module} can integrate the information of prototype patients with the health status of patients to obtain a deeper representation.

% \end{itemize}

\subsection{Multi-Channel Feature Extraction}
\label{sec:mc}
This module is developed to extract information from the input longitudinal patient matrix and static data to obtain the latent health status representations inspired by ConCare \cite{ma2020concare}. By embed each feature separately, the model can better model the changing pattern of each indicator.
% Though utilizing a single RNN is able to capture the feature interaction directly in each time step, embedding indicators separately can provide more detailed information on themselves and the interactions are performed in subsequent integration.
% We employ a multi-channel recurrent neural network. Each input feature corresponds to an RNN, and the embedding results are concatenated. Besides, we utilize a fully connected layer to embed this information due to the immutability of demographic data for each patient. In this way, the model can better model the changing pattern of each indicator.

Specifically, given a sequence of medical records along with a patient, the health status representation of each indicator $h_n = \mathrm{GRU}_n(r_n)$, where GRU (Gate Recurrent Unit) is a type of RNN (Recurrent Neural Network).
The demographic data embedded by a one-layer feed-forward network (FFN) can be denoted as $h_{static} = \mathrm{FFN}(r_{static})$. 
Then we concatenate the embedded laboratory tests and demographic data to acquire overall the patient's health status representation $h = [h_1,  h_2, ..., h_N, h_{static}], h \in \mathbb{R}^{(N+1) \times H}$, where $H$ denotes the dimension of latent space.

\subsection{Progressive Prototype Memory}
% In this section, we discuss our prototype memory used for interacting with patients. 
We select and store typical patients in memory as prototypes.
To obtain prototypes that are unrelated to outcomes, we cluster the latent representation $h$ of all patients in the training set to obtain $K$ clusters and take the patients closest to centroids as the prototypes $\mathbb{P} = \{p_1, p_2, ..., p_K\}$, where $p_j \in \mathbb{R}^{(N+1) \times H}$.
While training, the representation of patients in latent space may shift and decrease the interpretability, so we re-select prototypes at certain epochs progressively to ensure the cluster structure. The clustering algorithm is applied to calculate centroids. The nearest patients to each centroid in the latent space will be new prototypes to replace old prototypes via $p_j \leftarrow \mathop{\arg\min}\limits_{i} \Vert h^i-p_j\Vert_2$, where $i$ denotes the patient index.

Considering new and old prototypes may represent different subtypes of patients, and replacing them would significantly impact the upper layers, we match the old and new prototypes with the closest overall distance by Jonker-Volgenant algorithm \cite{crouse2016implementing}. After selecting prototypes, we freeze embedding layers and the prototype layer and then train the upper network for several epochs to guarantee stability. 

\subsection{Prototypical Feature Integration}
% Many previous works based on prototype learning make predictions merely via the similarity or distance between prototypes and the input. 
% However, we notice that this operation neglects the origin information from input and leads. 
For composing the learned prototypes and health status representation $h$, we create prototypical feature integration that can adaptively combine information of the given patients and prototypes while providing interpretation. Here, the similarity are only adopted to re-weight the health status representation to prevent neglecting of patient information.

% Firstly, we calculate the cosine similarity $s_{j}$ between patient and prototype $j$:

% \begin{equation}
% s_{j} = \frac{h \cdot p_j}{\left\|h\right\|_2 \left\|p_j\right\|_2}.
% \end{equation}

% In this module, the similarity $s$ will act as the query in the attention mechanism, and the health state $h$ of the patient is regarded as the key and value. 
In the integration operation, PPN first computes the similarity coefficient $s \in \mathbb{R}^{K}$ between the prototypes and patients via cosine similarity. The coefficient $s$ enhances prototypes with strong correlation and suppresses others.
% The query, key, value is computed by linear transformations $q = f_q(s), k = f_k(h^\top), v = f_v(h^\top)$, and $q \in \mathbb{R}^{H \times 1}, k \in \mathbb{R}^{H \times K}, v \in \mathbb{R}^{H \times K}$. These operation can also capture feature interactions.

% The query, key, value is computed by $q = W_q s, k = W_k h^\top,  v = W_v h^\top$, where $W_q \in \mathbb{R}^{K \times H}$, $W_k \in \mathbb{R}^{(N+1) \times K}$, and $W_v \in \mathbb{R}^{(N+1) \times K}$ are trainable matrices that can capture feature interactions. And the fusion weights $w = k^\top q, w \in \mathbb{R}^K$.

% \begin{equation}
% \begin{gathered}
% q = W_q s, \\
% k = W_k h^\top, \\
% v = W_v h^\top.
% \end{gathered}
% \end{equation}
% The attention weights $\alpha \in \mathbb{R}^K$ is denoted as follows:
\begin{equation}
s=\frac{ph}{\Vert p \Vert \times \Vert h \Vert}
\end{equation}

Then we calculate fusing health status $h_p$ associated with the coefficient $s$  by
\begin{equation}
h_p = h'^\top s,\quad h' = W_h h^\top,
\end{equation}
where $W_h \in \mathbb{R}^{(N+1) \times K}$. 

\subsection{Learning Objective}

With the integrated health status $h_p$, the prediction of health risks can be obtained by

\begin{equation}
\hat{y} = \sigma(f_{out}(h_p)).
\end{equation}
$\sigma$ represents the sigmoid operation, and $f_{out}$ is a one-layer feed-forward network. We use binary cross-entropy loss as the primary learning objective:

\begin{equation}
\mathcal{L}_c = -y\mathrm{log}\hat{y} - (1-y)\mathrm{log}(1-\hat{y}).
\end{equation}

Though progressive prototype selection update prototypes to ensure cluster structure in certain epochs, is not enough to choose patients from different subtypes. Therefore, we propose prototype separation losses to address this issue.
Since the prototypes can be regarded as particular patients, we predict the health risk $y_j$ of prototypes $j$ by $p_j$ as health status and adopt a separation loss that encourages the prototypes to represent different subtypes of patients:

\begin{equation}
\mathcal{L}_p = \sum^K_j \sum^{K}_{j' \ne j} y_j \mathrm{log} y'_j + (1 - y_j) \mathrm{log} (1 - y'_j).
\end{equation}

Nevertheless, one critical question that may arise is how to avoid trivial problems of learning the prototypes \cite{li2022prototype}.
For example, some prototypes may collapse into a single point in the latent space, or the prototype coefficients are assigned equally to some prototypes regardless of the input patient. 
These situations lead to reduced interpretability.
\cite{li2018deep} proposed two regularization terms to encourage a clustering structure in the latent space by minimizing the squared distance between an encoded sample and its closest prototype. But these terms may cause some samples closer to the cluster they do not belong to.
Consequently, we introduce another separation term $\mathcal{L}_d$:
\begin{equation}
\mathcal{L}_d = \sum^K_j \sum^{K}_{j' \ne j} \mathrm{max} (0, margin - \left\| p_j - p_{j'} \right\|_2).
\end{equation}

% $\mathcal{L}_p$ motivates prototypes represent patients of different health risk and $\mathcal{L}_d$ prevents prototypes collapse.
With loss terms formulated above, we can define the final objective of our model as

\begin{equation}
\mathcal{L} = \mathcal{L}_c + \lambda_p \mathcal{L}_p + \lambda_d \mathcal{L}_d,
\end{equation}
where $\lambda_p$ and $\lambda_d$ are hyper-parameters that control the excitation of prototype separation losses. We use a gradient descent back-propagation algorithm to minimize the loss $\mathcal{L}$ and optimize the parameters.

While training, the representations of prototypes may shift in latent space and they are not readily interpretable. Thus, we design an assigning step during training that assigns $p_j$ with the closest patient' embedding $h^i$ of patient $i$ in the training set: $p_j \leftarrow \mathop{\arg\min}\limits_{i} \left\|h^i - p_j\right\|_2$.

\section{Experiments}

In this section, we introduce the datasets and our experimental setup. Then we evaluate our model compared to the state-of-the-art baselines. We conduct analyses showing the effectiveness and interpretability of our proposed modules.

\subsection{Datasets}
We follow experiment settings from previous works \cite{ma2020concare,ma2020adacare,gao2020stagenet,zhang2021grasp}.
ESRD, Cardiology, and MIMIC-III datasets are applied for evaluation. The private information are desensitized when preprocessing.
ESRD dataset is an end-stage renal disease dataset, collected from diagnosis and treatment data of the peritoneal dialysis department of a hospital. Cardiology dataset \cite{reyna2019early} is collected from three U.S. ICUs for sepsis prediction. The goal of MIMIC-III dataset \cite{johnson2016mimic,harutyunyan2019multitask} is to predict in-hospital mortality based on the first 48 hours of an ICU stay. We illustrate the statistics of these datasets in Table \ref{dataset}.

\begin{table}[h]
  \setlength{\abovecaptionskip}{0pt}
  \setlength{\belowcaptionskip}{0pt}
  \setlength{\abovedisplayskip}{0pt}
  \setlength{\belowdisplayskip}{0pt}
  % \small
  \centering
  \begin{tabular}{llll}
    \toprule
     & ESRD & Cardiology & MIMIC-III \\
    \midrule
    Patient & 662 & 40,336 & 21,139 \\
    Positive outcomes & 258 & 2,932 & 2,797 \\
    Indicators & 17 & 34 & 17 \\
    Total Visits & 13,108 & 1,552,210 & 1,014,672 \\
    Maximum visits & 69 & 336 & 48 \\
    Minimum visits & 1 & 8 & 48 \\
    \bottomrule
  \end{tabular}
  \vspace{-6pt}
  \caption{Statistics of datasets}
  \label{dataset}
  \vspace{-12pt}
\end{table}

% To evaluate our model,  By comparing these two metrics, the predictive ability of PPN can be accurately evaluated.

\subsection{Experimental Setup and Baselines}
The prototypes are initialized by centroids at the beginning of the training.
% To reduce overfitting, the dropout \cite{srivastava2014dropout} method and layer normalization \cite{ba2016layer} are used. 
The hidden size $H$ of all modules is 32.
% The prototype replacing algorithm will be executed in epoch 10, 30 and 50.
% All models have trained 70 epochs with a batch size of 256, and the learning rate is $1 \times 10^{-3}$. 
We use mini-batch K-Means algorithm while computing centroids to accelerate. To reduce randomness, the old prototypes are regarded as initial centroids in the initialization of K-Means. We try different $K \in \{4, 6, 8, 12, 16, 20, 24, 28\}$ on these datasets, and a larger $K$ does not always lead to better results. Finally, when training ESRD dataset, we choose $K = 6, \lambda_p = 0.1$ and $\lambda_d = 0.05$. For Cardiology dataset, we set $K = 16, \lambda_p = 0.02$ and $\lambda_d = 0.01$. For MIMIC-III dataset, we set $K = 8, \lambda_p = 0.01$ and $\lambda_d = 0.005$. $margin = 70 / \sqrt{K}$ is utilized for all datasets. The progressive prototype selection will be executed in epochs 10, 30, and 50. The prototype assigning operation will be executed in every epoch. AUPRC (Area Under the Precision-Recall Curve)and AUROC (Area Under the Receiver Operating Characteristic Curve) are applied as evaluation indexes following \cite{ma2020adacare,ma2020concare,zhang2021grasp}.
% For a fair comparison with baseline models, their hyperparameters are fine-tuned through a grid search strategy. To evaluate precisely, we repeat every experiment 3 times with different random seeds and compute the standard variance.

We compare several state-of-the-art models as baselines: RETAIN \cite{choi2016retain}, 
SAnD \cite{song2018attend},
StageNet \cite{gao2020stagenet},
ConCare \cite{ma2020concare}, and
% MCGRU is a multi-channel GRUs model to embed time series data with an FFN to embed static data, which is similar to the structure in section \ref{sec:mc}.
GRASP \cite{zhang2021grasp}. ConCare is adopted as the backbone of GRASP.

% \begin{itemize}[leftmargin=*]
% \item Retain \cite{choi2016retain} utilizes two attention mechanisms to attend the EHR data in time order.

% \item SAnD \cite{song2018attend} employs self-attention, positional encoding, and dense interpolation to improve performance.

% \item AdaCare \cite{ma2020adacare} uses the dilated convolutions with multi-scale receptive fields and SE-block to capture the variation of biomarkers.

% \item StageNet \cite{gao2020stagenet} contains a stage-aware LSTM module and a stage-adaptive convolutional module to extract disease progression stage information.

% \item Grasp \cite{zhang2021grasp} discovers the patients with similar health status in their mini-batch, aggregates them into cohorts, and extracts representation to predict.
% \end{itemize}

\begin{table*}[h]
  \setlength{\abovecaptionskip}{0pt}
  \setlength{\belowcaptionskip}{0pt}
  \setlength{\abovedisplayskip}{0pt}
  \setlength{\belowdisplayskip}{0pt}
  %   \footnotesize
    \centering
    \begin{tabular}{llcccccc}
      \toprule
      & & \multicolumn{2}{c}{ESRD} & \multicolumn{2}{c}{Cardiology} & \multicolumn{2}{c}{MIMIC-III} \\
      \midrule
      & Model & AUPRC & AUROC & AUPRC & AUROC & AUPRC & AUROC \\
      \midrule
      % & GRU & 0.714(0.088) & 0.809(0.055) & 0.671(0.026) & 0.925(0.009) & 0.497(0.026) & 0.857(0.010) \\
      \multirow{6}{*}{Baseline} & RETAIN & 0.722$\pm$0.005 & 0.807$\pm$0.004 & 0.700$\pm$0.012 & 0.944$\pm$0.004 & 0.518$\pm$0.005 & 0.859$\pm$0.001 \\
      & SAnD & 0.658$\pm$0.003 & 0.767$\pm$0.012 & 0.302$\pm$0.015 & 0.764$\pm$0.021 & 0.477$\pm$0.013 & 0.848$\pm$0.002 \\
      & StageNet & 0.728$\pm$0.008 & 0.821$\pm$0.004 & 0.698$\pm$0.008 & 0.935$\pm$0.001 & 0.492$\pm$0.005 & 0.854$\pm$0.001\\
      & AdaCare & 0.722$\pm$0.007 & 0.811$\pm$0.005 & 0.710$\pm$0.022 & 0.939$\pm$0.008 & 0.493$\pm$0.004 & 0.855$\pm$0.001 \\
      & ConCare & 0.727$\pm$0.004 & 0.826$\pm$0.004 & 0.757$\pm$0.004 & 0.951$\pm$0.003 & 0.518$\pm$0.004 & 0.860$\pm$0.000 \\
      & GRASP & 0.733$\pm$0.005 & 0.830$\pm$0.003 & 0.774$\pm$0.002 & 0.955$\pm$0.002 & 0.525$\pm$0.002 & 0.863$\pm$0.001 \\
      \midrule
      \multirow{4}{*}{Reduced} & PPN$_{r-}$ & 0.749$\pm$0.011 & 0.838$\pm$0.008 & 0.777$\pm$0.002 & 0.961$\pm$0.001 & 0.536$\pm$0.008 & 0.867$\pm$0.001 \\
      & PPN$_{s-}$ & 0.755$\pm$0.002 & 0.843$\pm$0.003 & 0.785$\pm$0.006 & 0.960$\pm$0.001 & 0.531$\pm$0.005 & 0.865$\pm$0.001 \\
      & PPN$_{a-}$ & 0.671$\pm$0.015 & 0.765$\pm$0.008 & 0.784$\pm$0.002 & 0.959$\pm$0.005 & 0.473$\pm$0.008 & 0.849$\pm$0.002 \\
      & PPN$_{cluster}$ & 0.748$\pm$0.003 & 0.845$\pm$0.004 & 0.768$\pm$0.004 & 0.959$\pm$0.002 & 0.517$\pm$0.005 & 0.863$\pm$0.001 \\
      \midrule
      Proposed & PPN & \textbf{0.760$\pm$0.007} & \textbf{0.848$\pm$0.009} & \textbf{0.801$\pm$0.003} & \textbf{0.965$\pm$0.002} & \textbf{0.542$\pm$0.004} & \textbf{0.871$\pm$0.003} \\
      \bottomrule
    \end{tabular}
    \vspace{-6pt}
    \caption{Performance results on three dataset}
    \label{result}
    \vspace{-6pt}
\end{table*}

\subsection{Experiment Results}

We report the performance of PPN and other baseline models on two datasets in Table \ref{result}. PPN shows stable and outstanding performance and achieves state-of-the-art scores. Concretely, PPN achieves an average of 3.5\% higher AUPRC and 1.4\% AUROC than the best baseline model GRASP. 
SAnD cannot work well on ESRD and Cardiology dataset since it is designed for time series with a fixed length.

For further testing the performance of progressive prototype selection and prototype separation losses, we remove the progressive selection (PPN$_{r-}$) and the separation terms $\mathcal{L}_p$ and $\mathcal{L}_d$ (PPN$_{s-}$), respectively. Besides, the similarity coefficient $s$ is applied to predict instead of the prototypical feature integration to examine the power of the proposed module (PPN$_{a-}$). 
In Table \ref{result}, the reduced models still outperform baselines in most cases. Specifically, by comparing PPN and PPN$_{a-}$, we discover that predicting based on patients' representation and prototypes' information is significantly better than similarity alone.
Furthermore, we evaluate the cluster losses from \cite{ruis2021independent} instead of progressive selection and separation losses with prototypical feature integration in our framework (PPN$_{cluster}$), demonstrating the effectiveness of our approach. 

\subsection{Results on Missing Data}
\begin{figure}[h]
  \setlength{\abovecaptionskip}{0pt}
  \setlength{\belowcaptionskip}{0pt}
  \setlength{\abovedisplayskip}{0pt}
  \setlength{\belowdisplayskip}{0pt}
  \centering
  \subfigure[Different visit rate] {
  \label{visit}
  \includegraphics[width=0.47\columnwidth]{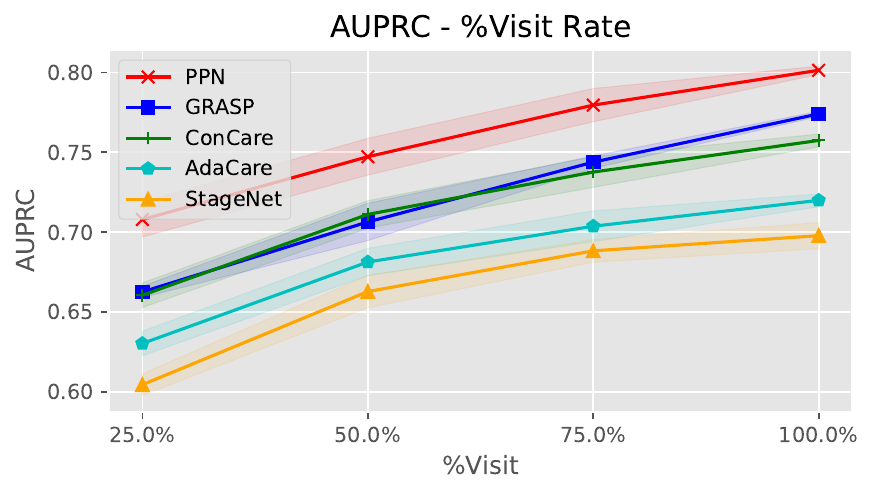}
  }
  \subfigure[Different observation rate] {
  \label{record}
  \includegraphics[width=0.47\columnwidth]{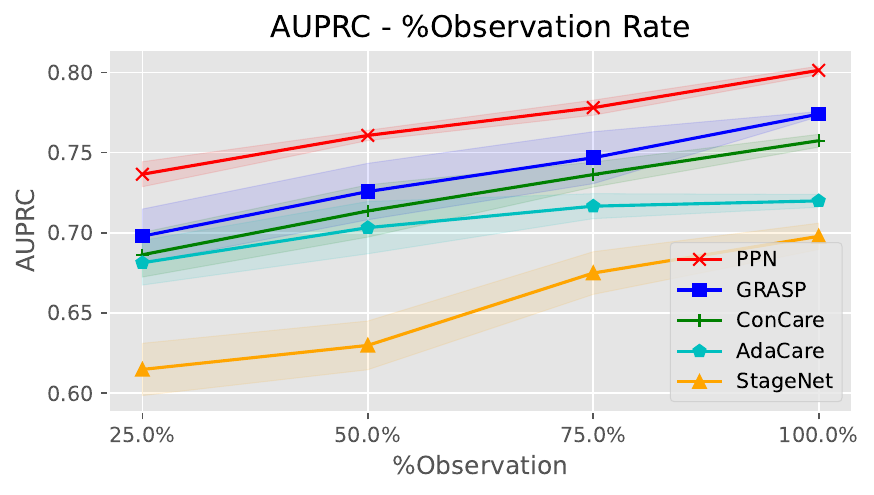} 
  }
  \vspace{-6pt}
  \caption{Experiments on different visit rate and observation rate. The shadow represents the standard variance}
  \vspace{-6pt}
\end{figure}

We evaluate whether PPN can utilize information from prototypes under patients with few visits or sparse features on the Cardiology dataset.
% Different ratio of visits (25\%/50\%/75\%/100\%) and observations (25\%/50\%/75\%/100\%) are examined. 
% The visits are randomly selected and observations are randomly masked. 
Figure \ref{visit} and Figure \ref{record} show the AUPRC in these scenarios, respectively.
% We can observe that PPN can handle these situations well and bring improvements. 
AdaCare, the CNN-based model, is not sensitive to the missing rate and performs worse. 
Compared to the RNN-based models (GRASP, ConCare, and StageNet), PPN performs better on low visit and observation rates, demonstrating that utilizing prototype information is fruitful. 
Although GRASP adopts similar patients to enhance representation, the similar patient is unreliable as the missing ratio increases. 
PPN achieves a 6.8\% improvement in the 25\% visit rate and a 5.5\% improvement in the 25\% observation rate compared to GRASP. 
Our model can adaptively select prototypes which are more credible and more robust. 

\subsection{Prototype Interpretations}
To interpret the predictions, we illustrates the information about typical patients selected by PPN in Table \ref{prototypical}. We found that each typical patient came from a different population, from young to old, male or female, diabetic or non-diabetic, showing their diversity.

For interpreting in the perspective of cohorts, we create a cohort set $\mathbb{S}_j$ for each typical patient $j$ containing all the patients whose most similar typical patient is $j$. Formally, $\mathbb{S}_j = \{i | s_{ij} > s_{ij'}, \forall j' \neq j\}$, where $s_{ij}$ is the similarity between patient $i$ and typical patient $p_j$.
% $S_j = \mathop{\arg\max}\limits_{i} s_{i, j}$. 
Then we calculate the average age, male ratio, mortality rate, and diabetes rate in Table \ref{statictics}. The statistics among $\mathbb{S}_j$s are different, further confirming the diversity of prototypes learned by PPN. The similarity between patients and typical patients can be exploited for individual interpretation. We visualize the progression of the disease with the trajectory of changing cohorts of patients in the application so that physicians can understand the predictions.

% \begin{table}[h]
%   \caption{Statistics about average age, male ratio, mortality rate, diabetes rate, and distinctive features (ranked by p-value) of $S_j$ on the ESRD dataset}
%   \label{statictics}
%   \centering
%   \setlength{\tabcolsep}{3px}{
%   \begin{tabular}{cccccccc}
%     \toprule
%     Index & Avg. Age & Male Ratio & Mortality Rate & Diabetes Rate & \multicolumn{3}{c}{Distinctive Features}\\
%     \midrule 
%     \#0 & 63.71 & 0.962 & 0.423 & 0.683 & Weight & Glucose & Hb\\
%     \#1 & 58.31 & 0.144 & 0.351 & 0.333 & Albumin & Weight & CO$\mathrm{_2}$CP\\
%     \#2 & 38.74 & 0.116 & 0.095 & 0.098 & Weight & DBP & Glucose\\
%     \#3 & 44.03 & 0.947 & 0.182 & 0.221 & Creatine & Weight & Appetite\\
%     \#4 & 74.95 & 0.783 & 0.761 & 0.457 & Albumin & DBP & Creatine\\
%     \#5 & 72.76 & 0.070 & 0.563 & 0.414 & Creatine & Weight & Appetite\\
%     \bottomrule
%   \end{tabular}
%   }
% \end{table}

\begin{table}[h]
  % \vspace{-3pt}
  \setlength{\abovecaptionskip}{0pt}
  \setlength{\belowcaptionskip}{0pt}
  \setlength{\abovedisplayskip}{0pt}
  \setlength{\belowdisplayskip}{0pt}
  \centering
  \begin{tabular}{cccccc}
    \toprule
    Index & Age & Gender & Outcome & Diabetes & PD\\
    \midrule 
    \#0 & 58.93 & Male & Positive & True & DN\\
    \#1 & 62.07 & Female & Negative & False & CIN\\
    \#2 & 42.28 & Female & Positive & False & CGN\\
    \#3 & 44.94 & Male & Positive & False & CGN\\
    \#4 & 79.44 & Male & Negative & True & DN\\
    \#5 & 81.26 & Female & Negative & True & DN\\
    \bottomrule
  \end{tabular}
  \vspace{-6pt}
  \caption{Information of typical patients on the ESRD dataset. PD: primary disease. DN: diabetic nephropathy. CIN: chronic interstitial nephritis. CGN: chronic glomerulus nephritis.}
  \vspace{-6pt}
  \label{prototypical}
\end{table}

\begin{table}[h]
  \setlength{\abovecaptionskip}{0pt}
  \setlength{\belowcaptionskip}{0pt}
  \setlength{\abovedisplayskip}{0pt}
  \setlength{\belowdisplayskip}{0pt}
  \centering
%   \small
%   \setlength{\tabcolsep}{0.9mm}{
  \begin{tabular}{ccccc}
    \toprule
    Index & Avg. Age & Male Ratio & Mortality & Diabetes\\
    \midrule 
    \#0 & 38.74 & 11.6\% & 9.5\% & 9.8\%\\
    \#1 & 44.03 & 94.7\% & 18.2\% & 22.1\%\\
    \#2 & 58.31 & 14.4\% & 35.1\% & 33.3\%\\
    \#3 & 63.71 & 96.2\% & 42.3\% & 68.3\%\\
    \#4 & 72.76 & 7.0\% & 56.3\% & 41.4\%\\
    \#5 & 74.95 & 78.3\% & 76.1\% & 45.7\%\\
    \bottomrule
  \end{tabular}
  \vspace{-6pt}
  \caption{Statistics of $\mathbb{S}_j$ on the ESRD dataset}
  \label{statictics}
  \vspace{-6pt}
\end{table}

% \begin{table}[h]
%   \centering
%   \begin{tabular}{cccc}
%     \toprule
%     Index & \multicolumn{3}{c}{Distinctive Features} \\
%     \midrule 
%     \#0 & Weight & DBP & Glucose\\
%     \#1 & Creatine & Weight & Appetite\\
%     \#2 & Albumin & Weight & CO$\mathrm{_2}$CP\\
%     \#3 & Weight & Glucose & Hb\\
%     \#4 & Creatine & Weight & Appetite\\
%     \#5 & Albumin & DBP & Creatine\\
%     \bottomrule
%   \end{tabular}
%   \caption{Most distinctive features (ranked by p-value) of $\mathbb{S}_j$}
%   \label{statictics_t}
% \end{table}

% In order to identify distinctive features of $\mathbb{S}_j$, we apply the T-test and find 3-10 significant features in each set (p-value $\le$ 0.05). The top 3 significant features ranked by p-value are reported in Table \ref{statictics_t}. We can see that the distinctive features in low-risk sets (\#0 and \#1), medium-risk sets (\#2 and \#3), and high-risk sets (\#4 and \#5) are different, which indicates that PPN clearly distinguishes patients of different subtypes, even though their health status is similar. Creatine, weight, and appetite are essential biomarkers of patients’ nutritional status \cite{yeun1998factors,post2019creatine}.
% % \cite{yeun1998factors}
% Albumin is crucial for evaluating the fundamental health condition \cite{bal2013binding}.
% Glucose is directly related to diabetes \cite{lawrence1940renal}. DBP is a decisive indicator for heart diseases such as cardiovascular \cite{wang2013reversed}, and heart diseases may be a complication of chronic kidney disease \cite{sarnak2003cardiovascular}.

\section{Conclusion}

In this work, we propose a progressive prototypical network to select and incorporate information from typical patients. 
% PPN consists of multi-channel feature extraction, progressive prototype memory, prototype feature integration, and health risk prediction modules.
PPN adopts progressively selection and two separation losses to learn prototypes from the dataset while ensuring the cluster structure and diversity. We also design a prototypical feature integration to utilize their information to enhance the representation for the given patient. Experiments on three real-world datasets prove that PPN consistently outperforms state-of-the-art methods. Additional experiment demonstrate the effectiveness of PPN when handling patients with few visits and sparse records. Moreover, we develop an interactive application to illustrate the interpretable predictions for physicians.
% The patients with sparse records may influence the accuracy of similarity, so we plan to compute uncertainty to fill this gap.
We hope our model can help physicians analyze patients through typical cases to diminish adverse outcomes.

\textbf{Acknowledgement:} This work was supported by the National Natural Science Foundation of China (No.82241052).

% References should be produced using the bibtex program from suitable
% BiBTeX files (here: strings, refs, manuals). The IEEEbib.bst bibliography
% style file from IEEE produces unsorted bibliography list.
% -------------------------------------------------------------------------
\bibliographystyle{IEEEbib}
\bibliography{reference}

\end{document}